\documentclass{article}
\PassOptionsToPackage{numbers}{natbib}
\usepackage[final]{nips_2017}

\usepackage[utf8]{inputenc} 
\usepackage[T1]{fontenc}    
\usepackage{hyperref}       
\usepackage{url}            
\usepackage{booktabs}       
\usepackage{amsfonts}       
\usepackage{nicefrac}       
\usepackage{microtype}      

\usepackage{graphicx}
\usepackage{bm}
\usepackage{amssymb}
\usepackage{amsmath}
\usepackage{mathtools}
\usepackage[tight]{subfigure}

\title{GANosaic: Mosaic Creation with Generative Texture Manifolds}

\author{
  Nikolay Jetchev\\
  \begin{footnotesize}
  \texttt{nikolay.jetchev@zalando.de} \end{footnotesize}\\
  Zalando Research\\  
  \And Urs Bergmann\\
  \begin{footnotesize}\texttt{urs.bergmann@zalando.de}\end{footnotesize} \\
  Zalando Research\\
  \And Calvin Seward \\
  \begin{footnotesize}
  \texttt{calvin.seward@zalando.de} \end{footnotesize}\\
  Zalando Research\\
}
\begin{document}
\maketitle
\vspace*{-0.5cm}

\begin{abstract}
  This paper presents a novel framework for generating texture mosaics with convolutional neural networks. 
Our method is called GANosaic and performs optimization in the latent noise space of a generative texture model, which allows the transformation of a content image into a mosaic exhibiting the visual properties of the underlying texture manifold. 
To represent that manifold, we use a state-of-the-art generative adversarial method for texture synthesis \cite{PSGAN2017}, which can learn expressive texture representations from data and produce mosaic images with very high resolution. 
This fully convolutional model generates smooth (without any visible borders) mosaic images which morph and blend different textures locally.
In addition, we develop a new type of differentiable statistical regularization appropriate for optimization over the prior noise space of the PSGAN model.

\end{abstract}

\section{Introduction}
\subsection{Mosaics and textures}
Mosaics are a classical art form. The Romans were masters in skillfully selecting small colored stones to make beautiful mosaics of large scenes. Later, in his paintings, the Renaissance painter Archimboldo composed various objects to make amazing portraits of people. In general, mosaics work because of the properties of the human visual system to average colors over spatial regions -- when looking from a distance the large image emerges, but when looking closely the details of the single tiles emerge. In modern times, computer graphic algorithms have enabled different forms of digital image mosaics~\cite{photomosaic,JIM}.  However, these methods use distinct non-overlapping small tiles to paint the large image. A seamless mosaic style like Archimboldo's -- where the whole image acts as a mosaic, without any tiles with borders -- is visually closer to modern methods of texture synthesis and transfer.

Image quilting~\cite{EfrosQ} recombines patches from the original textures in order to smoothly reconstruct a target image -- ``texture transfer". However, a disadvantage is the high runtime complexity when generating large images. In addition, since instance models merely copy the original pixels, they cannot be used to generalize and create novel textures from multiple examples.

The work of~\cite{GatysEB15a} uses discriminatively trained deep neural network as effective parametric image descriptors, allowing both texture synthesis and a novel form of texture transfer called ``neural art style transfer." However, texture synthesis and transfer is performed from a single example image and lacks  the ability to represent and morph textures defined by several different images.


Spatial versions of Generative Adversarial Networks (GAN) are well suited to unsupervised learning of textures~\cite{SGAN2016,PSGAN2017}. The Periodic Spatial GAN (PSGAN) allows high quality texture synthesis, with efficient memory and speed usage. It can also leverage information from many input images and use them to learn a texture manifold, a rich distribution over many textures allowing morphing into novel textures. Such generative models can give more widely varied outputs than the instance and neural descriptor based approaches to texture synthesis. Such variety is key to our proposed method.


\subsection{Introducing a new algorithm for mosaic creation: GANosaic}

\begin{figure}[t]
\centering
\includegraphics[width=9.7cm]{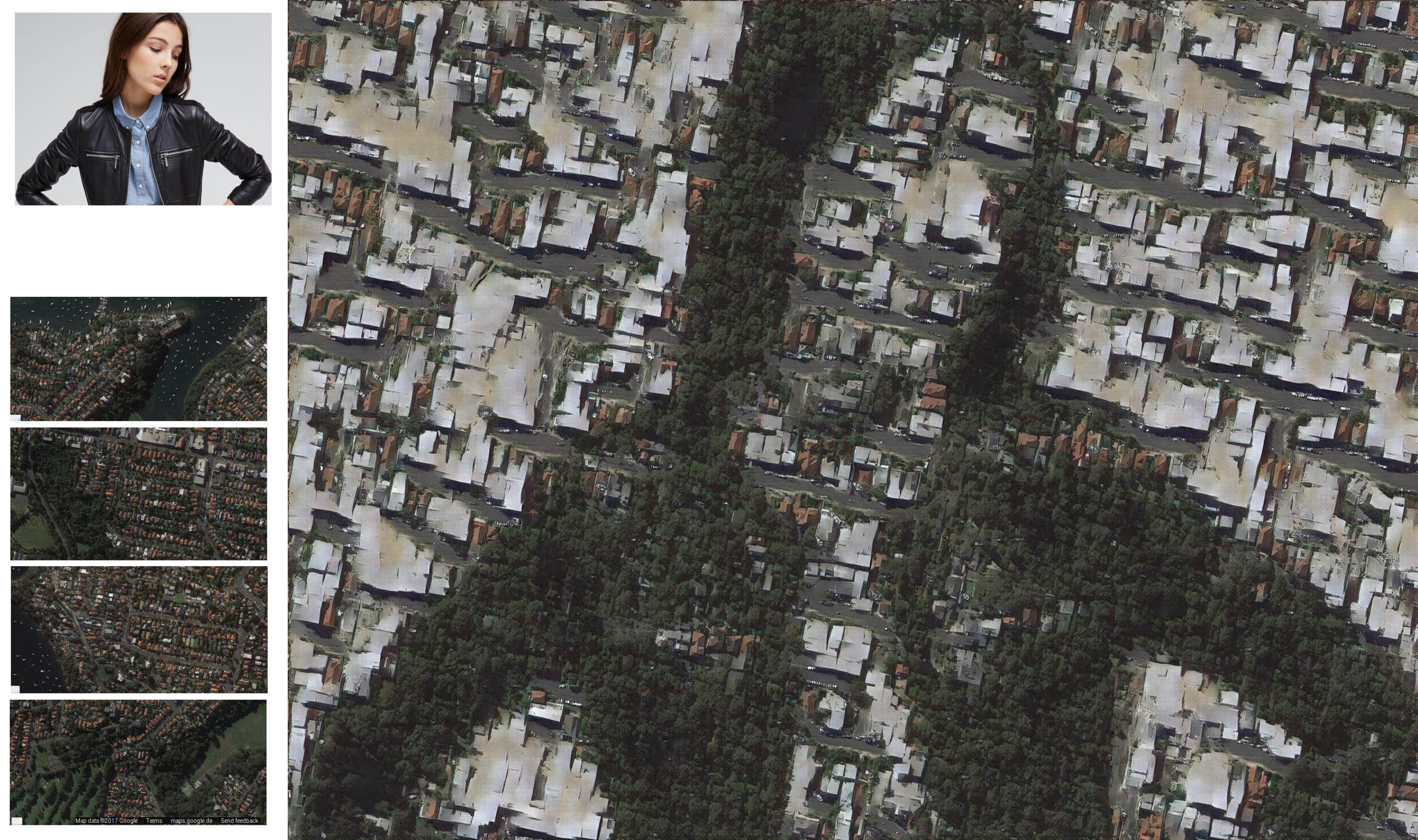} 
\caption{A texture manifold was learned with a PSGAN from satellite images of the city of Sydney, serving as texture prior for our neural mosaic. 
GANosaic renders a human portrait in a mosaic of size 1984x1472 pixels by minimizing a loss function -- paint the target image and stay close to the texture manifold --  w.r.t. the texture generator input (noise) tensor, with 62x46 spatial dimensions and 60 channels. The perceptual distance in the loss is represented by layer Conv5-1 of the pre-trained VGG~\cite{Simonyan14c} network. The mosaic is best seen zoomed-in to appreciate the rich small details of both city  and forest. The texture style images (shown scaled smaller on the left) had size 1770x860 pixels when used for training the PSGAN.}
\label{fig_sat}
\end{figure}

Our novel GANosaic method has two steps. First, a PSGAN is trained on a set of example images, for details see Section \ref{sec_psgan}. Second, the generator $G$ of the PSGAN is used as a module in an optimization problem: generate an image that is as close as possible to a ``content" image $I$, while staying on the learned texture manifold, the ``style". 
One way to do this is to adapt the input noise tensor $Z$ s.t. the output image from the generator $G(Z)$ is as close as possible to the target image $I$. This is done by defining the distance between $G(Z)$ and $I$ as a loss function and then optimizing it w.r.t. $Z$. However, during PSGAN training $Z$ came from a prior noise distribution, and the mosaic optimization can push $Z$ to values far away from the prior distribution and lead to degenerate looking textures.



For aesthetically pleasing mosaics we want $G(Z)$ to stay as close as possible to the texture manifold ``style". When we use a PSGAN texture model, the mosaic style is better represented with a texture loss term that ensures that the input $Z$ tensor\footnote{In our notation $Z$ denotes both the noise random variable and a tensor sampled from it.} stays close to its prior statistics used during PSGAN training. 
Concretely, we model the loss to ensure that the local channels $Z^l$ keep their statistical independence.
The total loss function is therefore composed of two parts, a content loss and a texture loss:


\begin{equation}
\mathcal{L}(Z) = \mathcal{L}_{c}(Z)  + \alpha_l   \mathcal{L}_{tex}(Z^l) \label{eq_loss}
\end{equation}

The term $\mathcal{L}_{c}$ denotes the content loss ensuring reconstruction of $I$ by $G(Z)$:
\begin{equation}
\mathcal{L}_{c}(Z) = \|\phi(I) - \phi(G(Z))   \|_F^2
\end{equation}

Here  $\| . \|_F^2$ denotes the mean of all squared tensor elements. 
The mapping $\phi$ is the ``correspondence map"~\cite{EfrosQ}, which specifies what perceptual distance metric we want to use w.r.t. the content image. This can be a simple predefined image transformation, or a more complex approach, e.g. utilizing the outputs of pretrained convolutional filters~\cite{GatysEB15a}.
By using some image downscaling operator in $\phi$ (e.g. pooling layers) we split the frequencies of the resultant image: the low frequencies are determined by the content image, and the high frequencies come from the texture manifold. Such a split improves the mosaic quality, 
see Section \ref{sec_ablation} for ablation studies regarding the effects of the choice of $\phi$. 

The texture loss is also required to keep the optimized noise tensor close to the manifold of textures created by the prior noise distribution. It regularizes the local noise channels $Z^l$ of the $Z$ noise tensor $Z = [Z^g,Z^l,Z^p]$. See Section \ref{sec_locreg} for more details on the loss $\mathcal{L}_{tex}$ and its effect on mosaic output. 

In summary, the GANosaic is a powerful novel method to generate art, with the following key properties:
\begin{itemize}
\item generation of seamless mosaics with unique texture visual aesthetics 
\item flexible differentiable texture model \citep{PSGAN2017} that learns and morphs diverse texture images 
\item very large scalability with respect to output mosaic size -- all calls to the generator $G(Z)$ can be efficiently split into small tensor chunks seamlessly forming a very large final image \citep{SGAN2016}
\item fast optimization in $Z$ noise space by gradient descent
\item exploration of multiple different mosaics for given texture and content image
\end{itemize}
Please see Figure \ref{fig_sat} for an example of mosaics that can be created by our method. Section \ref{sec_discuss} has additional discussion of the properties of GANosaic.


\section{Background: texture generation via PSGAN}
\label{sec_psgan}
This section contains a brief summary of the texture model PSGAN, for elaboration see~\cite{PSGAN2017}. 
Generative Adversarial Networks~\cite{Goodfellow14} learn a generator network $G(\bm{z})$ to distort a noise vector $\bm{z}$, which is sampled from a standard distribution (e.g. uniform), such that the distorted probability distribution is close to the distribution observed through the training samples of the form $X \in \mathbb{R}^{H \times W \times 3}$. This is achieved with a game theoretic idea, by letting the generator network $G$ play against an additional network, the discriminator $D$: the task of the discriminator is to classify a sample as being from the generator or from the training set, while the generator tries to be as good as possible in producing samples that get classified by the discriminator as real training data.

The extensions of PSGAN beyond the standard GAN framework are threefold. First, as in spatial GANs~\cite{SGAN2016}, the architecture is chosen to be a fully convolutional version of DCGAN~\cite{RadfordMC15} and the noise vector $\bm{z}$ is extended to a spatial tensor $Z \in \mathbb{R}^{L \times M \times d}$. Here, $L$ and $M$ are spatial dimensions, while $d$ is the channel dimension. Hence, akin to DCGAN, the fractionally strided convolutions in PSGANs upsample the spatial dimensions $L$ and $M$ to the output dimensions $H$ and $W$. In our case we typically use 5 convolution layers with each a fractional stride of $\frac{1}{2}$, hence the total upsampling in our case is $\frac{H}{L} = \frac{W}{M} = 2^{5} = 32$.

As the discriminator is chosen symmetrically to $G$, in particular also fully convolutional, the standard GAN cost function needs to be marginalized over the spatial classifications:
\begin{align}
V(D,G) =& \frac{1}{LM} \sum_{\lambda=1}^L \sum_{\mu=1}^M \mathbb{E}_{Z \sim p_Z(Z)} \left[ \log \left( 1 - D_{\lambda\mu}(G(Z)) \right) \right]  \nonumber \\
  +& \frac{1}{LM} \sum_{\lambda=1}^{L} \sum_{\mu=1}^{M} \mathbb{E}_{X' \sim p_{\mathrm{data}}(X)} \left[ \log D_{\lambda\mu}(X') \right],
  \label{eq:pgan_vfunc}
\end{align}
where $D_{\lambda\mu}(X)$ is the discriminator output at location $\lambda$ and $\mu$.
The key advantage of this approach is that the image patches used for the training minibatches can have different size than the image outputs used when sampling from the model, yielding arbitrary large output resolution. However, as the receptive field of a single location in the $Z$ tensor is spatially limited in the output, far away regions are independently sampled. The local statistics must therefore be independent of the position -- in other words, only sampling of a single texture is possible. To overcome this limitation, as a second extension, a fraction of channels in $Z$ are spatially shared to allow for conditioning on a global structure. 
The final extension is to implement a spatial basis in parts of $Z$, which can be used to anchor image generation. It has been shown in~\cite{PSGAN2017} that a plane wave parameterization for the spatial basis allows the generation of periodic textures, and can also lead to better quality of non-periodic textures. The wave numbers of the plane waves in PSGAN are given as functions of the global channels by a multi-layered perceptron, which is learned end-to-end alongside $G$ and $D$. 
In total the tensor $Z= [Z^g,Z^l,Z^p]$ consists of three parts: a local part $Z^l$, a global part $Z^g$, and a periodic part $Z^p$, concatenated in the channel dimension.

After learning, the spatially shared global channels define a texture to be sampled, while the independent samples in the local dimensions give rise to local pattern variation. Importantly, when the global channels are allowed to change smoothly in the spatial dimensions, this yields a spatial transitioning of textures, while being locally still plausible textures. Hence we speak of a texture manifold.

\begin{figure*}[tb]
 \center{
  \includegraphics[width=0.61\textwidth]{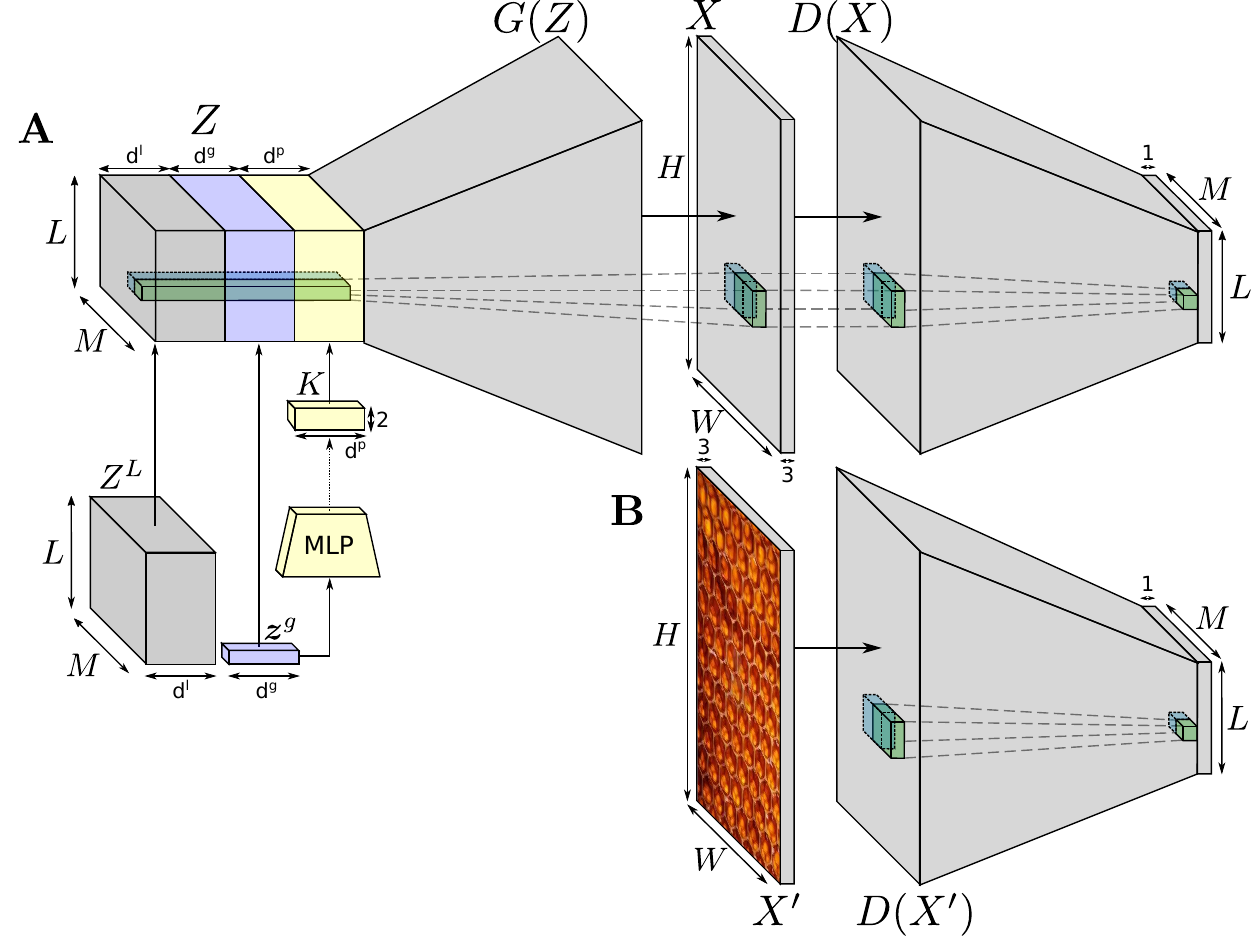} 
 }
 \caption{Illustration of the PSGAN model. \textbf{A} The fully convolutional generator network $G(Z)$ maps a spatial tensor $Z_{\lambda \mu i}$, $\lambda$ and $\mu$ being the spatial indices, to an input image $X$. Every subvector at a spatial location in $Z$, e.g.\ the blue or green columns in the Figure, map to a limited area in $X$. As usual in GAN training, the discriminator gets either a generated image $X$ or, as in \textbf{B}, an image patch $X'$ from the real data.}
  \label{fig:PSGAN}
\end{figure*}

\begin{figure}
\centering
\subfigure[Sydney]{\includegraphics[width=3.1cm]{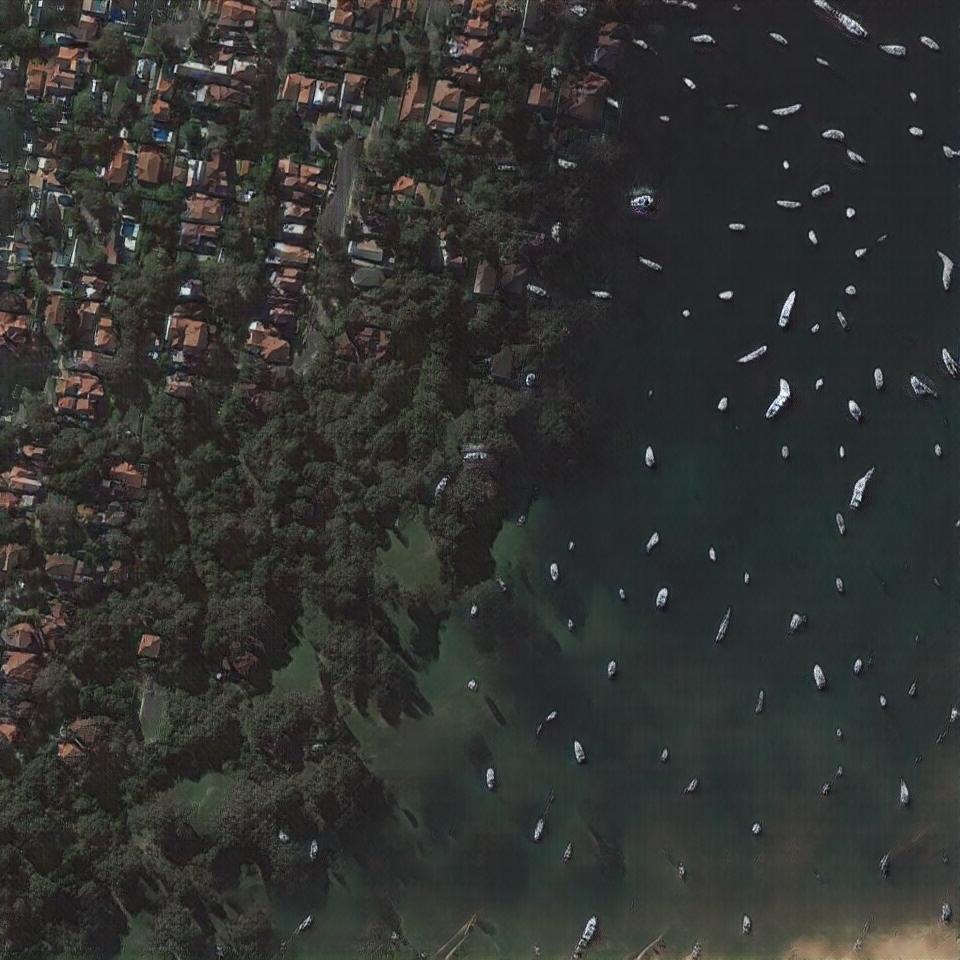}} 
\subfigure[stone walls]{\includegraphics[width=3.1cm]{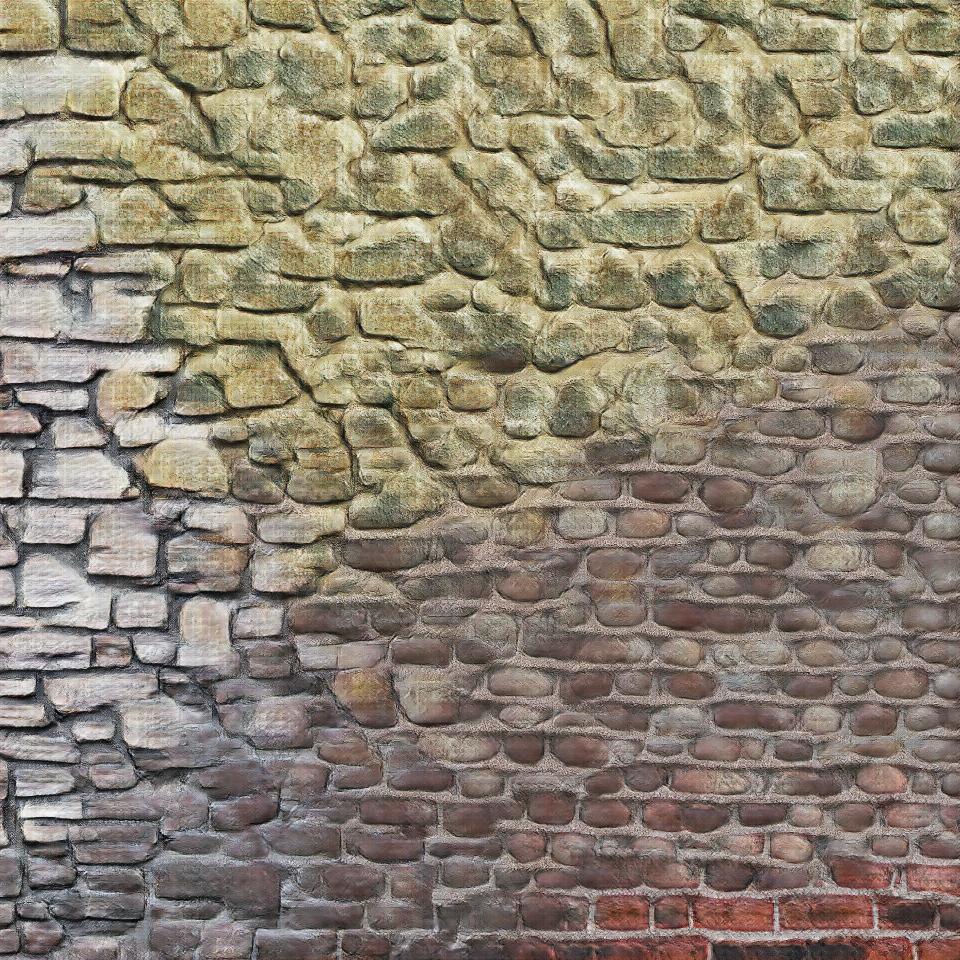} }
\subfigure[DTD ``scaly'']{\includegraphics[width=3.1cm]{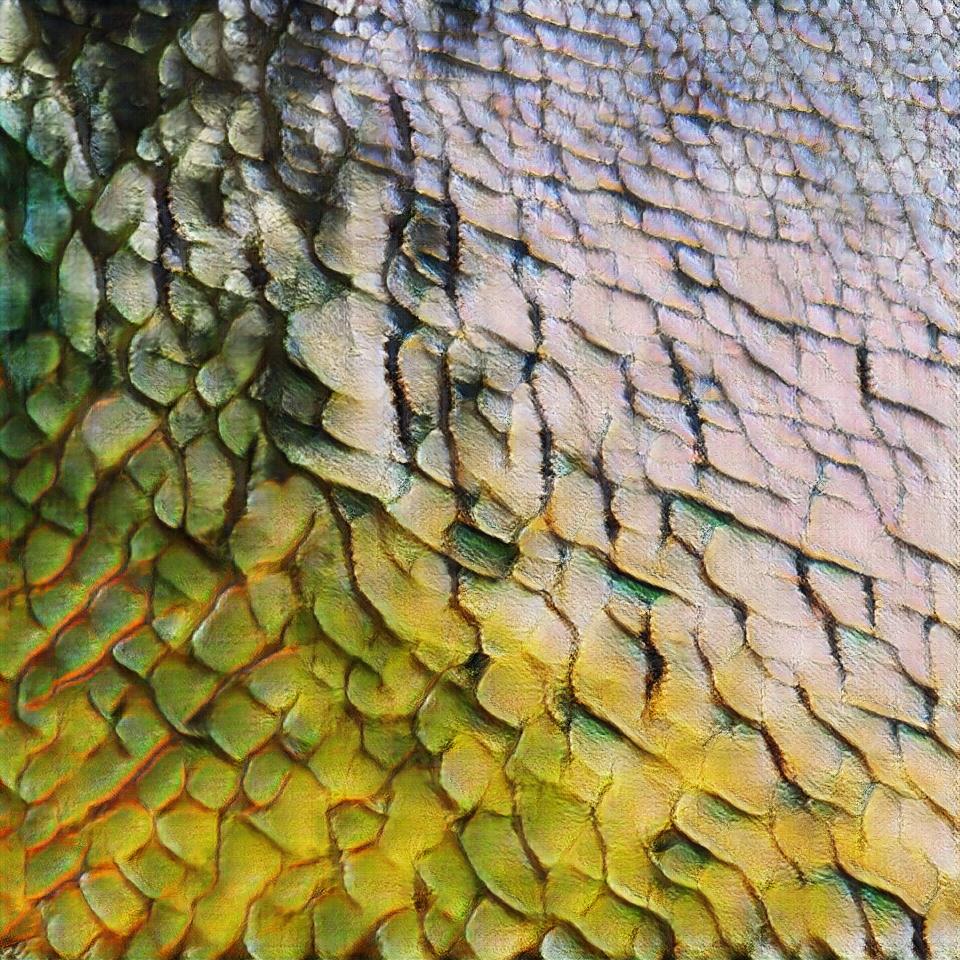} }
\caption{Examples of the PSGAN texture models used for our mosaics. Morph plots show the ability of texture manifolds to smoothly change texture processes. The plots (size 960x960 pixels) created by bi-linearly interpolating the $Z^g$ tensor (size $L=M=30$) between 4 random texture samples.}
\label{fig_text}
\end{figure}

Figure \ref{fig_text} shows how the texture manifolds learned by PSGAN look. We use input texture images for PSGAN training of Sydney satellite images from Google Maps, stone wall images from Wikimedia Commons, and DTD ``scaly" from ~\cite{cimpoi14describing}.

\subsection*{Technical note: fixed constant batch normalization in $G$ for GANosaic}
The generator $G$ of PSGAN was trained using batch normalization after every deconvolutional layer. 
The batch normalization calculates per layer statistics that capture the distribution of feature activations in the minibatch input to $G$. 
However, the GANosaic method (optimization of $Z$ w.r.t. a content and texture loss, $Z$ is a tensor with 1 instance of a large spatial extent) is different than the PSGAN training (optimization of $G$ parameters w.r.t. GAN loss function, $Z$ has many instances of spatially small arrays from the noise prior distribution). We found empirically that GANosaic  works better by using fixed statistics (mean and variance) for the batch normalization operations of the trained $G$. 
Concretely, we pre-calculate the batch normalization statistics of a  batch with many instances of $Z$ sampled from the prior. Afterwards,  these statistics are used as a constant rescaling for each batch normalization operation inside the network $G$. 
This also makes easier the practical implementation of splitting procedures \citep{SGAN2016} for very large mosaics, a key ability of GANosaic.

\section{Texture loss}
\label{sec_locreg}

Optimization of $Z^l$ w.r.t. the content loss $\mathcal{L}_c$ can introduce spatial correlations between the local dimensions. During training of the texture model, however, the local dimensions $Z^l$ in the PSGAN model were independently sampled from the prior at every spatial position and channel dimension. Hence, the correlations imply a move away from the learned texture manifold. To remedy this, we introduce a regularization term: the key idea is that samples taken from the joint distribution of neighboring local dimensions should be distributed according to the prior distribution during training (up to finite sample size effects). In our case, as the prior is an independent uniform distribution, this means the samples should fill up the whole hypercube $[0,1]^{L \times M}$. In contrast, if local dimensions were perfectly correlated, the samples would lie exclusively on the diagonal of the hypercube.

To implement this idea we assume independence in the channel dimension and considered the different channels as the samples. By employing a kernel density estimate, the joint distribution can be estimated and compared to the prior distribution. For practical reasons, only pairwise neighboring positions in $Z^l$ are evaluated. The restriction to neighboring positions can be justified by noting that correlations in natural images fall off monotonously with distance~\cite{hyvarinen2009natural}. The computational benefit is a reduction of quadratic to linear time complexity. 
Formally, we can write the texture loss term $\mathcal{L}_{tex}$ as:
\begin{equation}
\mathcal{L}_{tex}(Z^l) = \sum_{\lambda=1,\mu=1}^{L-1,M-1} 
\sum_{\Delta\lambda,\Delta\mu \in \Delta} d\left( \hat p_{\left[ Z^l_{\lambda,\mu},Z^l_{\lambda+\Delta\lambda,\mu + \Delta\mu} \right]}, \hat p_{\mathrm{prior}}\right)
\end{equation}
where $d(p_1, p_2)$ measures the distance of two probability distributions $p_1$ and $p_2$, and the square brackets denote the concatenation of column vectors to a matrix. The set of spatial offsets $\Delta$ determines for which neighboring positions the distribution is regularized. We took $\Delta = \{(0,1),(1,1),(1,0)\}$.

The kernel density estimate given $\hat Z \in \mathbb{R}^{d_l \times 2}$ and evaluated at a point $\bm{\tau}$ is given as $\hat p_{\hat Z}(\bm{\tau}) = \frac{1}{d_l \sigma}\sum_{i=1}^{d_l} k(\frac{\| \left[ \hat Z_{i1}, \hat Z_{i2} \right] - \bm{\tau} \|^2}{\sigma})$. Any valid kernel function can be used for $k$; we employed a Gaussian kernel.
From the form of $\hat p_{\hat Z}$ we see that the target probability distribution of the regularizer is the convolution of the prior probability distribution with the Gaussian kernel, i.e. $\hat p_{\mathrm{prior}} = p_{\mathrm{prior}} * k$. 

\begin{figure}[t]
\centering
\subfigure[Reference density $\hat p_{\mathrm{prior}}$.]{\includegraphics[width=5cm]{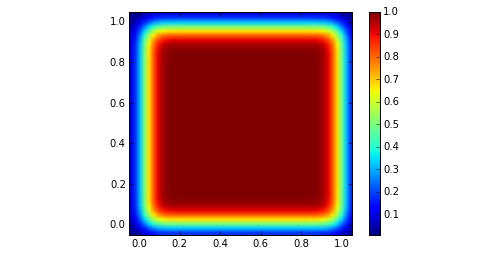} }
\subfigure[The optimized blue points are more consistent with the prior.]{\includegraphics[width=4.5cm]{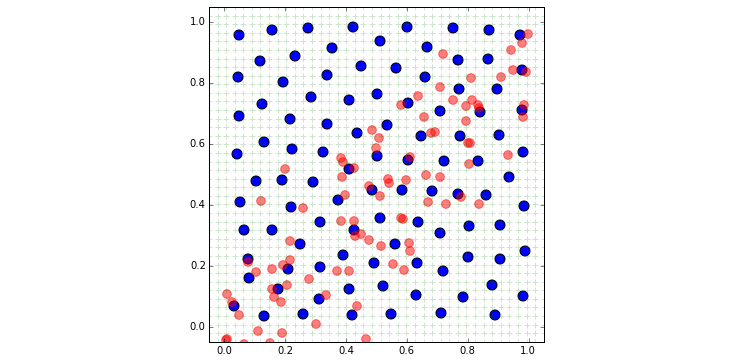} }
\caption{ (a) The reference density $\hat p_{\mathrm{prior}} = p_{\mathrm{prior}} * k$, using the uniform distribution on $[0,1]\times[0,1]$ as prior $p_{\mathrm{prior}} $, and a Gauss kernel $k$ with $\sigma=0.1$ for density estimation. (b) $100$ samples (red dots) are drawn from a different distribution than the prior. We use a grid with $N=40 \times 40$ regularly spaced points (green crosses) to measure the distance of the reference density and the kernel density estimate of the samples. Minimizing the distance yields samples (blue dots), which are more consistent with the prior $p_{\mathrm{prior}}$ than the initial samples (red).} \label{fig_kde}
\end{figure}

Finally, the distance function $d$ between the probability distributions needs to be defined. We simply calculate the distance as the square difference between the two distributions evaluated on the set $\mathcal{G}$ of grid points equally spaced in the unit cube, i.e. $d(p_1, p_2) = \sum_{\bm{\tau} \in \mathcal{G}} \| p_1(\bm{\tau}) - p_2(\bm{\tau}) \|^2$. This makes the regularizer a differentiable function w.r.t. $Z$. 
Figure~\ref{fig_kde} gives a toy example of the behavior of the regularizer. Note that the regularizer is similar to determinantal point processes~\cite{kulesza2012determinantal}, in particular the resulting samples tend to be too regular in comparison to samples from the prior.

\section{Experiments}

\subsection{Experimental setup}\label{sec_bfgs}
For the texture optimization procedure we used gradient optimization (BFGS from SciPy), and we constrained all $Z$ values to lie in $[-1,1]$, to be compatible with the prior distribution. 
The speed for a single gradient step is 0.2 seconds for a 1024x768 pixel image, 0.4 seconds for 1600x1000 pixels, on a Tesla GPU K80. Usually less than 20 iterations are enough to get nice looking mosaic images. 
These timings apply for a simple perceptual distance map $\phi$ and our standard PSGAN architecture (see below) -- using a neural network for $\phi$ can cost more time, depending on the layers and channels. In general, we expect that a very large content image and a very rich PSGAN model (trained on many textures) can make the optimization more complicated and require more iterations to optimize, but we did not inspect this in detail. With small texture sets of a few training images we got expressive PSGAN models that could be used for good-looking mosaics.

While our optimizer worked quite well when starting from a single texture (same $z^g$ on every spatial position), we found that doing a few iterations of stochastic search (via sampling random projections, see Section \ref{sec_rproj}) and using the sampled mosaic with lowest loss as initialization can help the optimizer reach solutions slightly better than a trivial initialization from a random texture. In addition, by randomly sampling initializations for the optimizer, we are able to explore multiple different solutions to the mosaic loss optimization, which is another source of aesthetic variety.

\subsection{Effects of the content loss correspondence map}
\label{sec_ablation}
\begin{figure}[tb]
\centering
\subfigure[No downscaling]{\includegraphics[width=5.5cm]{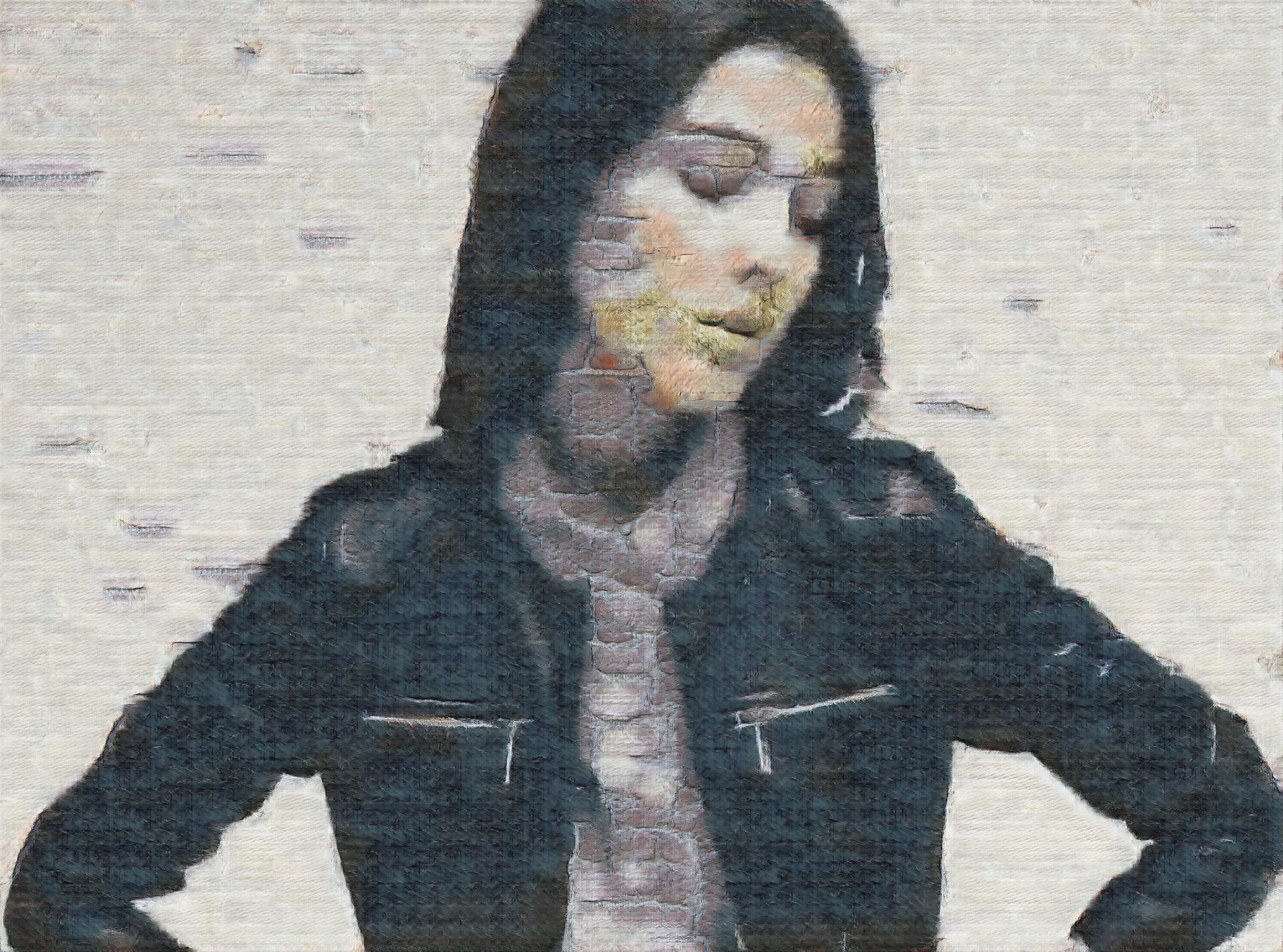} }
\subfigure[Downscale 4x]{\includegraphics[width=5.5cm]{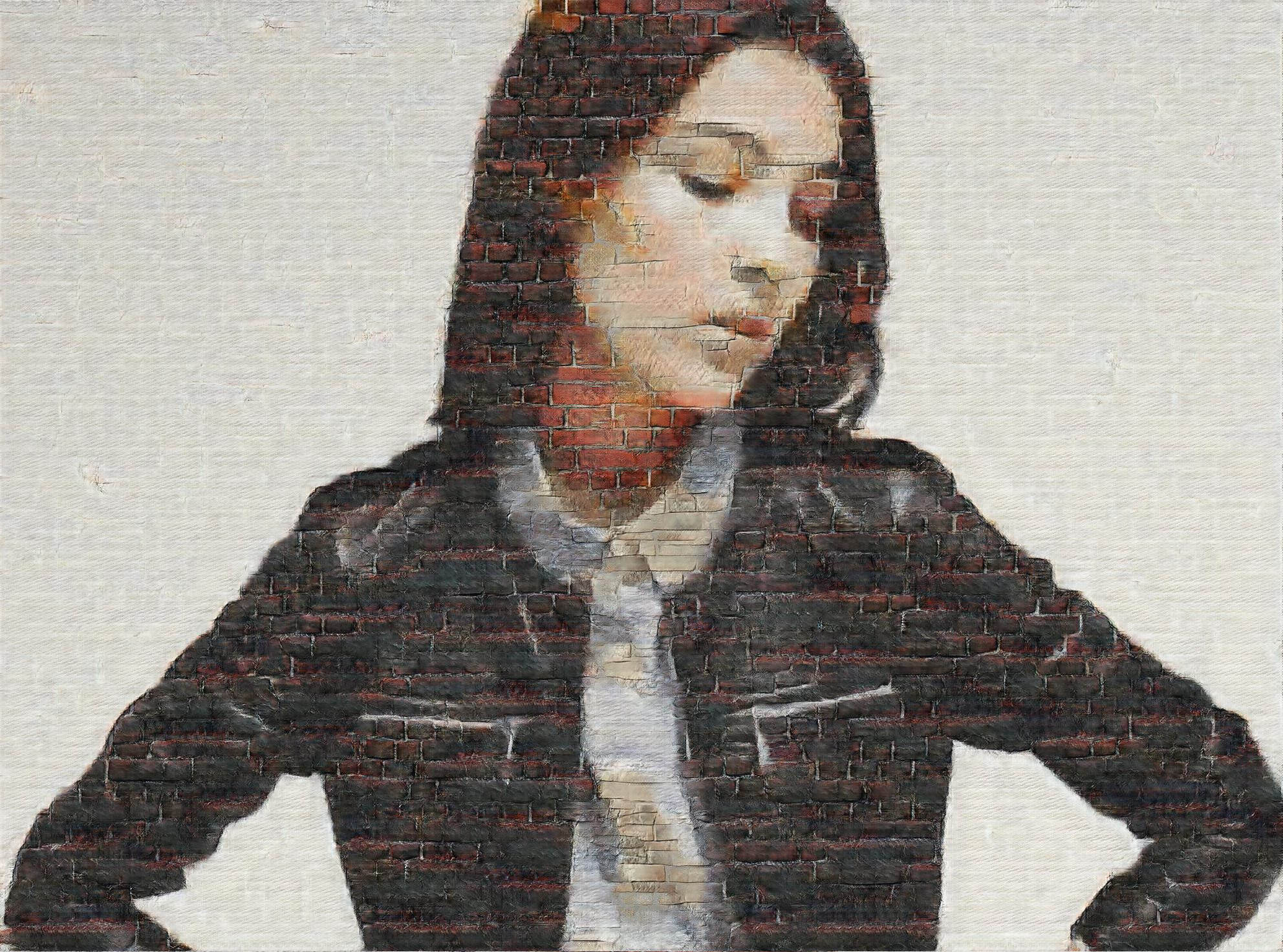} }
\subfigure[Downscale 16x]{\includegraphics[width=5.5cm]{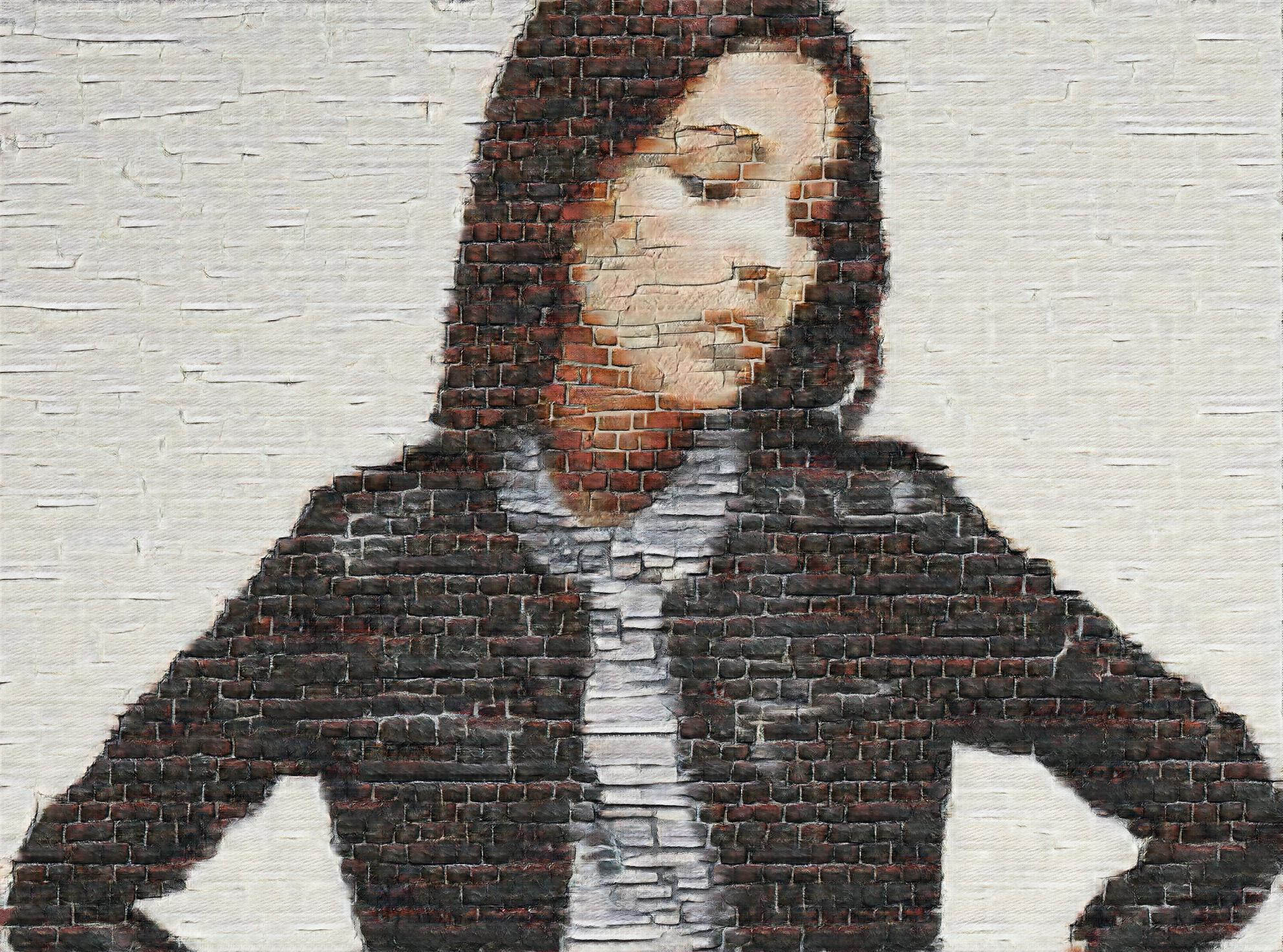} }
\subfigure[Downscale 64x]{\includegraphics[width=5.5cm]{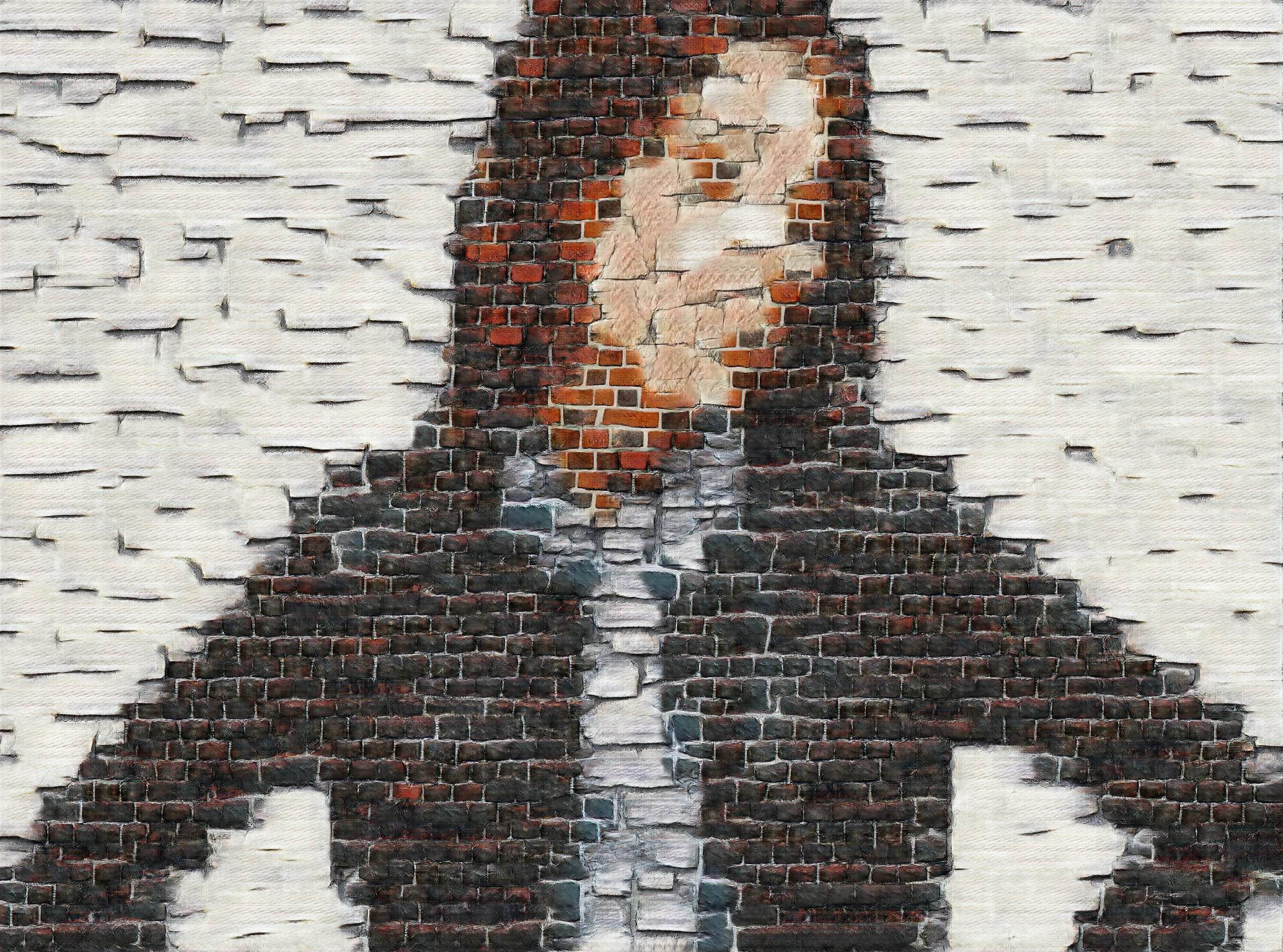} }
\caption{Downscaling by average pooling in the correspondence map $\phi$  allows a tradeoff in the content loss: accurate content image rendering (a) or strong texture aesthetics (d). Mosaics (b) and (c) are more balanced. All 4 mosaics have the resolution 1984x1472 pixels and use 7 images of stones from Wikimedia Commons for textures.}
\label{fig_dsample}
\end{figure}

The correspondence map $\phi$ used for the content loss encodes directly a choice of perceptual distance metric -- it allows us to have flexibility in the transfer of the texture appearance on the mosaic. If we use the identity as map, then for some images the mosaic output will be degenerate. Figure \ref{fig_dsample}(a) shows this drawback of the identity correspondence map when the content image has too high frequencies, which are difficult to map to the texture manifold. Adding downscaling (e.g. with an average pooling filter) to  $\phi$ will emphasize the lower frequencies in the content image and lead to better texture appearance, see Figure\ref{fig_dsample}(b,c). Downscaling too much will make the mosaic reproduce the content image less accurately as in Figure \ref{fig_dsample}(d), but as a trade-off the stone textures are really well recognizable.

\begin{figure}
\centering
\subfigure[Luminance, Downscale 4x]{\includegraphics[height=4.4cm]{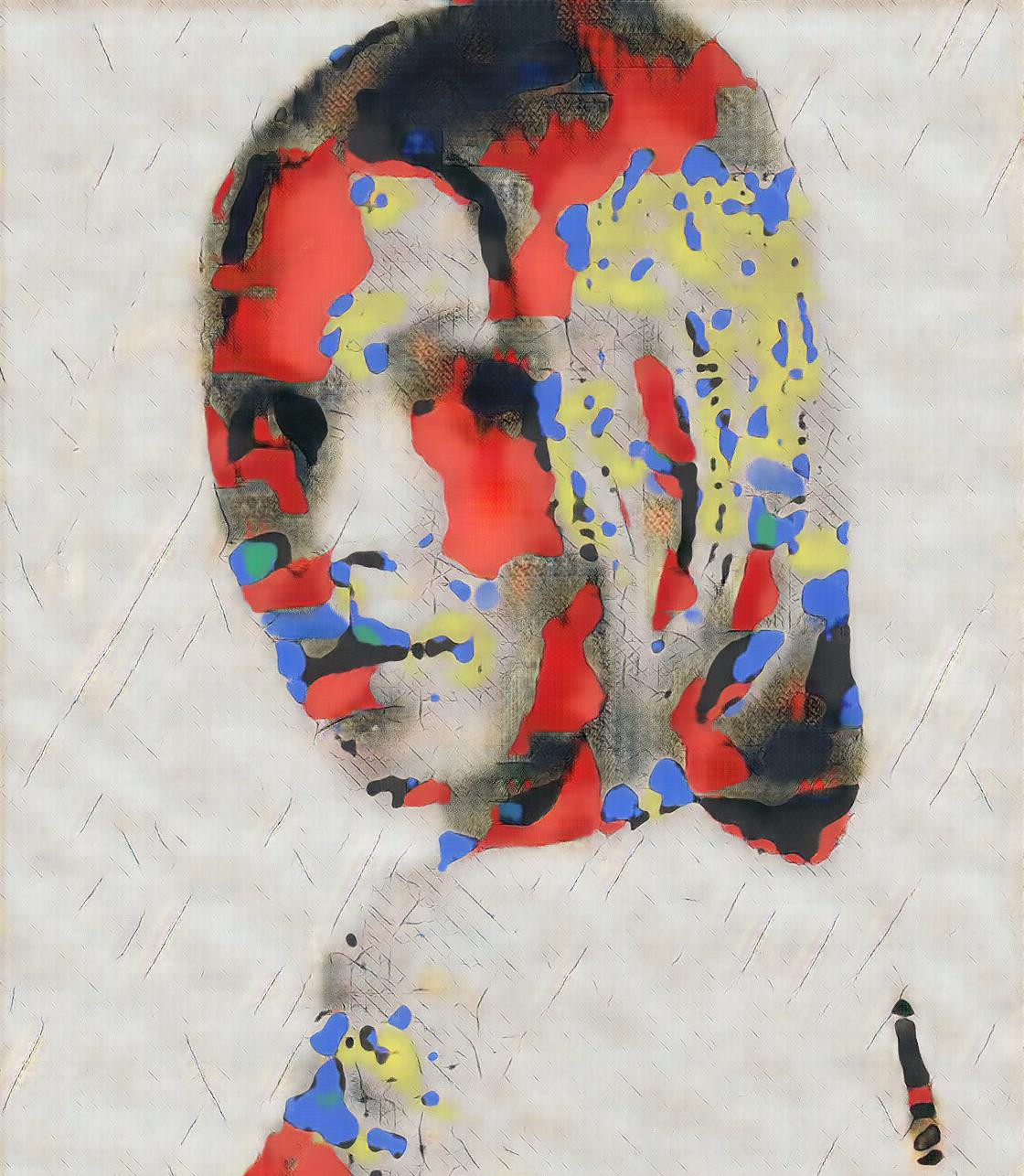}}
\subfigure[\text{Conv4-2}]{\includegraphics[height=4.4cm]{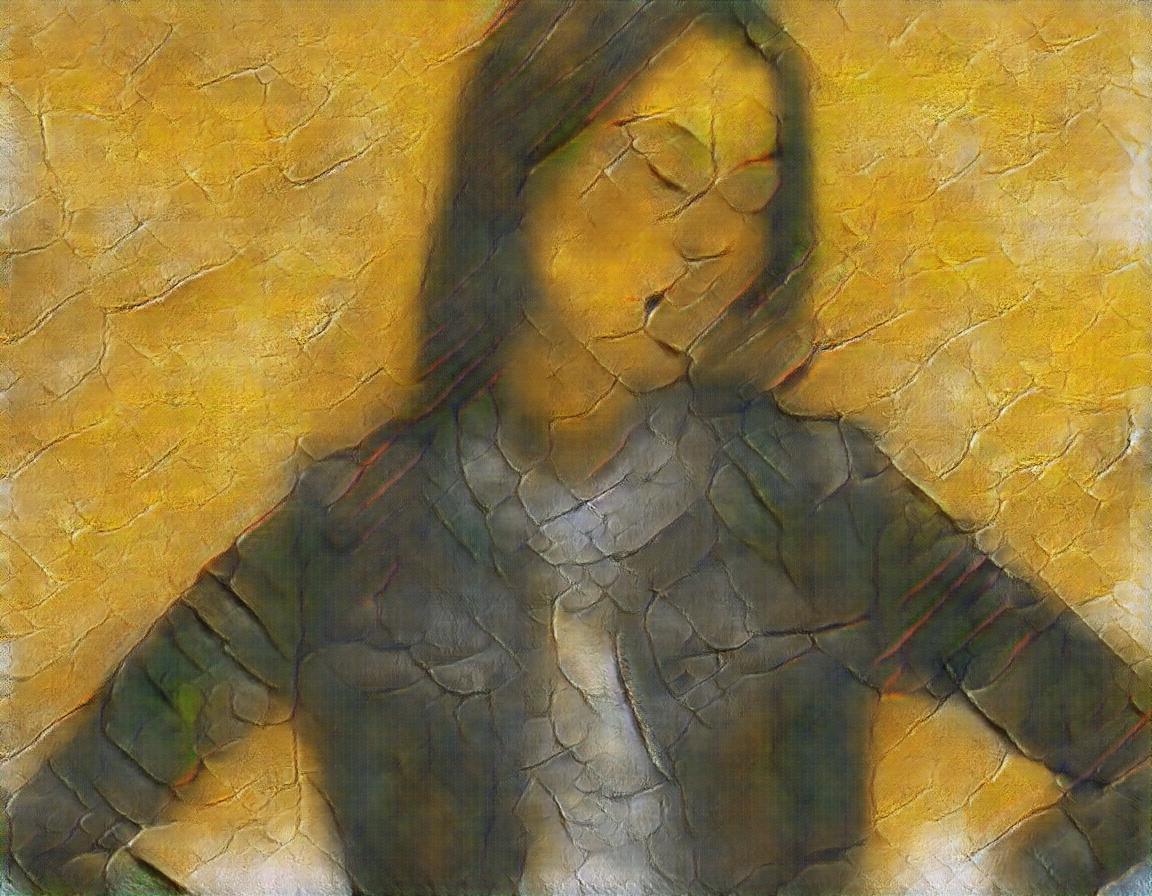}}
\subfigure[RGB, Downscale 16x]{\includegraphics[height=4.4cm]{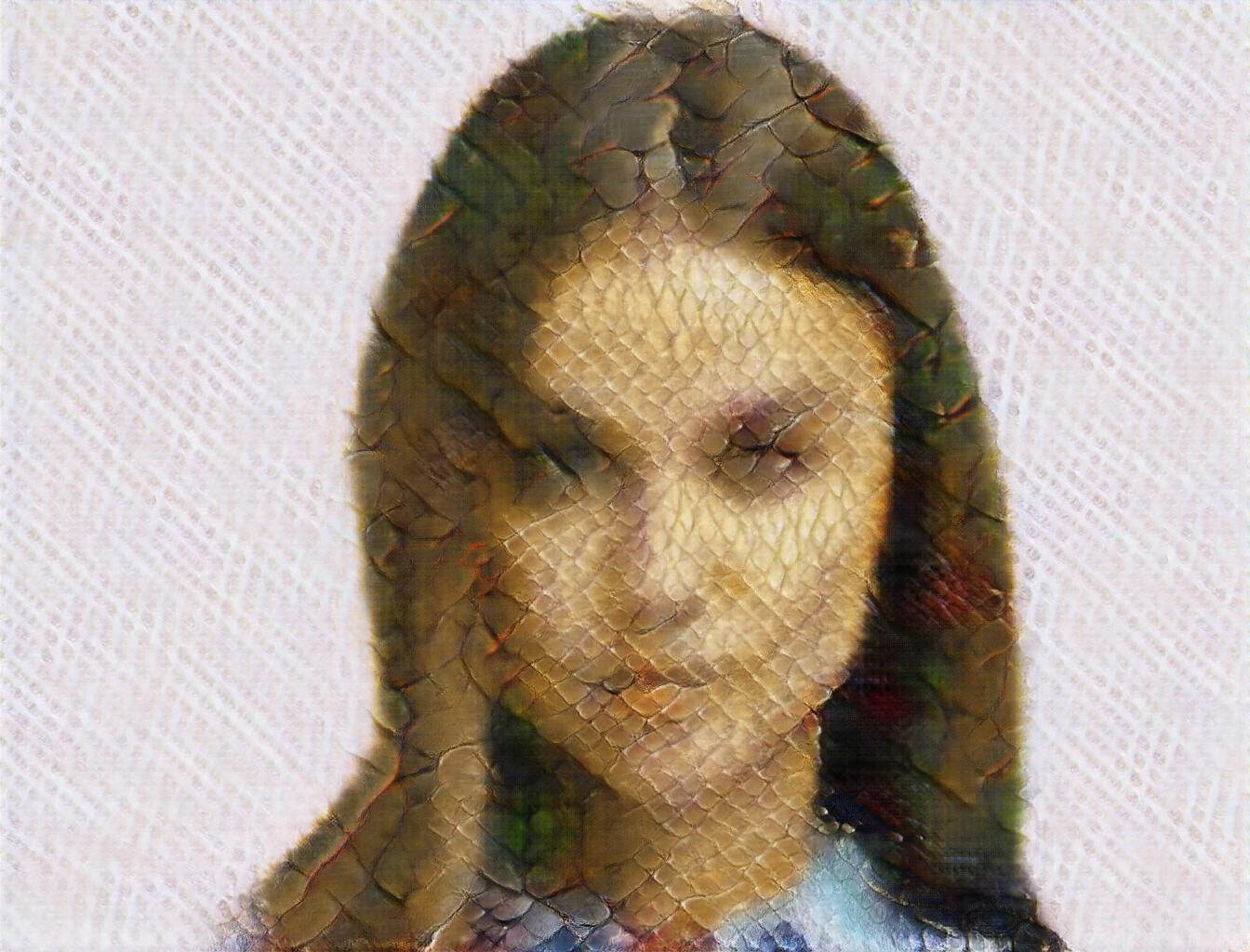}}
\subfigure[RGB, Down. 4x]{\includegraphics[height=4.4cm]{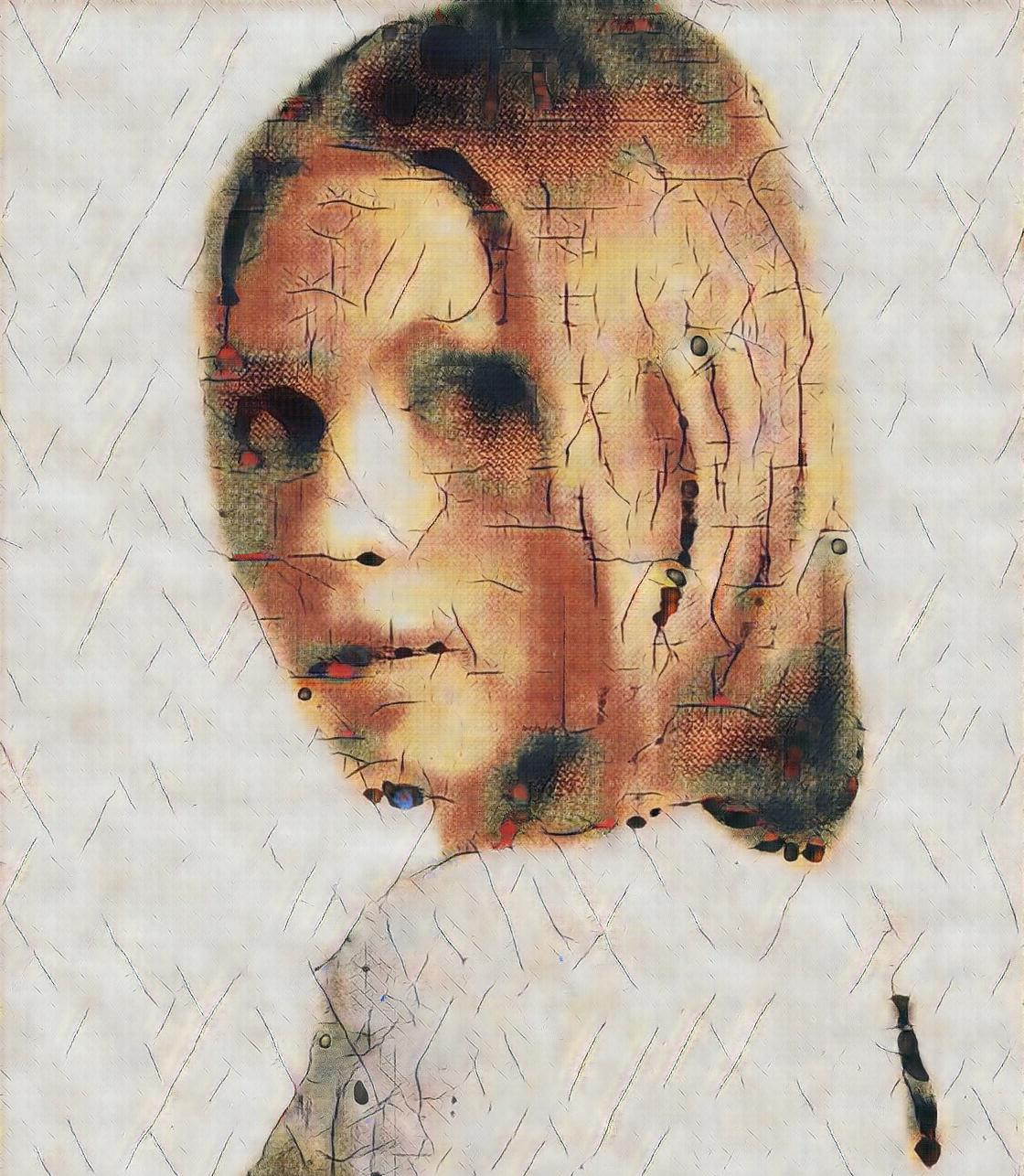}}
\subfigure[\text{Conv5-1}]{\includegraphics[height=4.4cm]{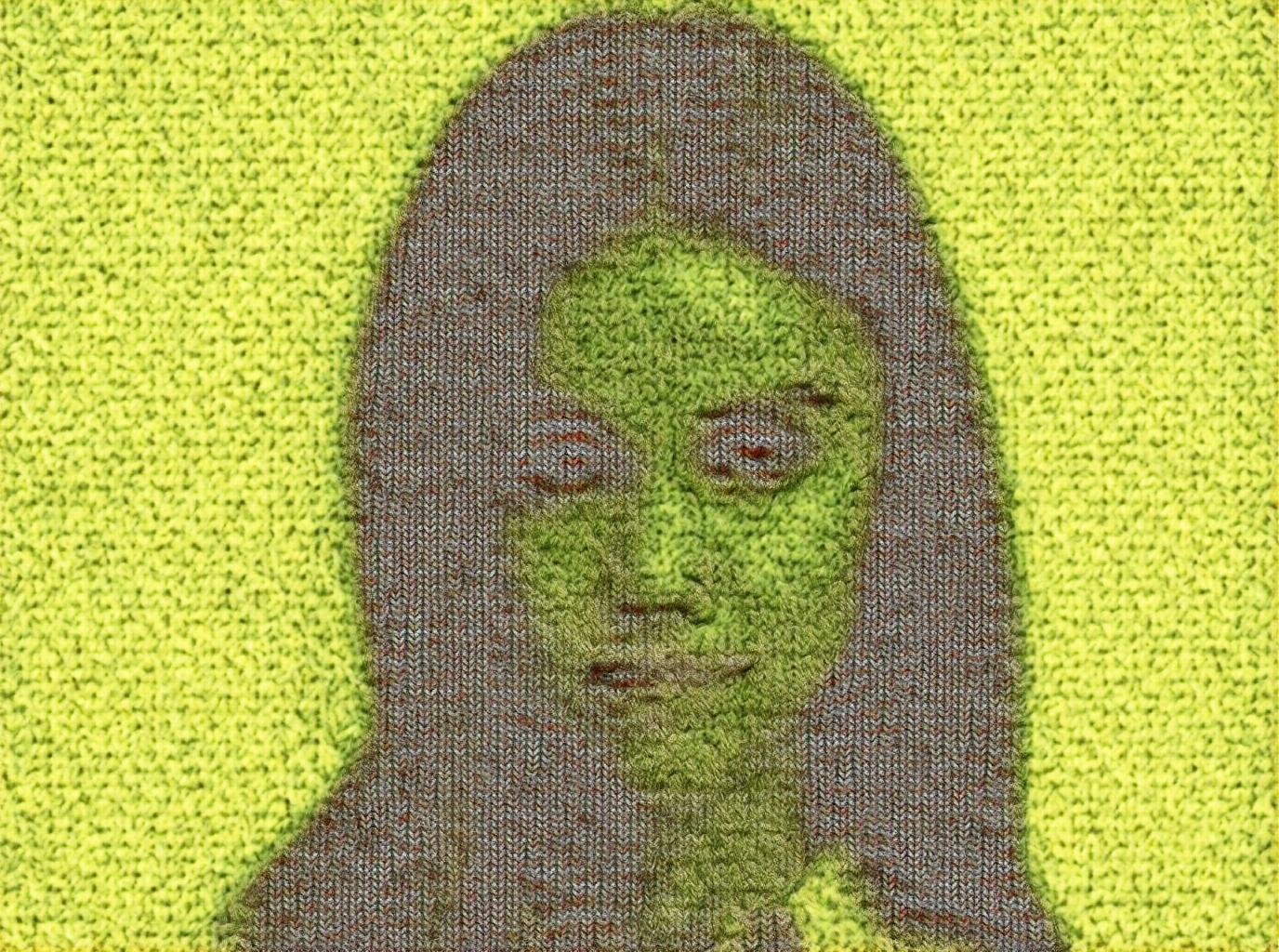}}
\subfigure[\text{Conv5-1}]{\includegraphics[height=4.4cm]{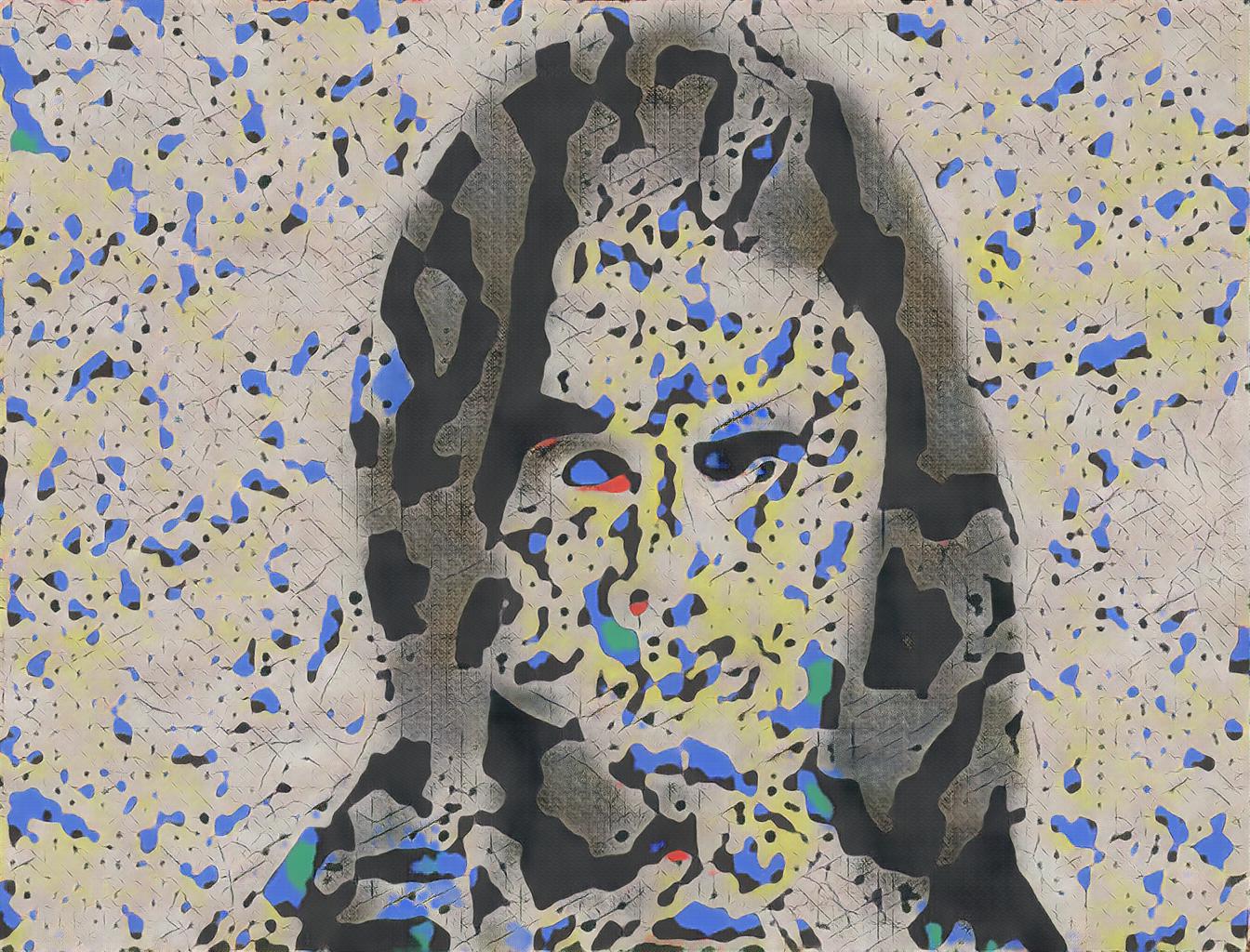}}
\caption{Results using different  correspondence maps. (a) shows how using only the luminance allows to make mosaics with a more experimental look, away from the original color palette as used in (d). In (b,e,f) layers of the VGG network encode more complex perceptual distances. Used textures: DTD ``knitted" in (e), DTD ``scaly" in (b,c), Juan Miro paintings in (a,d,f). Please zoom-in for details.}
\label{fig_vgg}
\end{figure}

Another interesting choice for $\phi$ may be to use instead of the exact colors the luminance channel of the RGB image, defined by us as the average of the 3 RGB channels. This is useful when the texture manifold is very different in color hue from the content image. Figure \ref{fig_vgg}(a) shows an example mosaic with the luminance map, and it has more color variation than the RGB map \ref{fig_vgg}(d).

The downscaling and color transformations are examples of manually specified correspondence maps $\phi$. As an alternative, we can take filters from the pretrained VGG network~\cite{Simonyan14c}, which is similar to the approach of~\cite{GatysEB15a}.
Figure \ref{fig_vgg}(b,e,f) shows our results with VGG, which has a different aesthetics than the other choices of $\phi$ -- this perceptual distance is more flexible w.r.t. color hue and also is more flexible w.r.t. spatial matching of the content. This is due to the VGG network architecture, which is deep (many convolutional layers) and wide (many filter channels) and uses pooling operations.

\subsubsection{Disabling the mosaic content loss}
\label{sec_rproj}

\begin{figure}
\centering
\includegraphics[width=8cm]{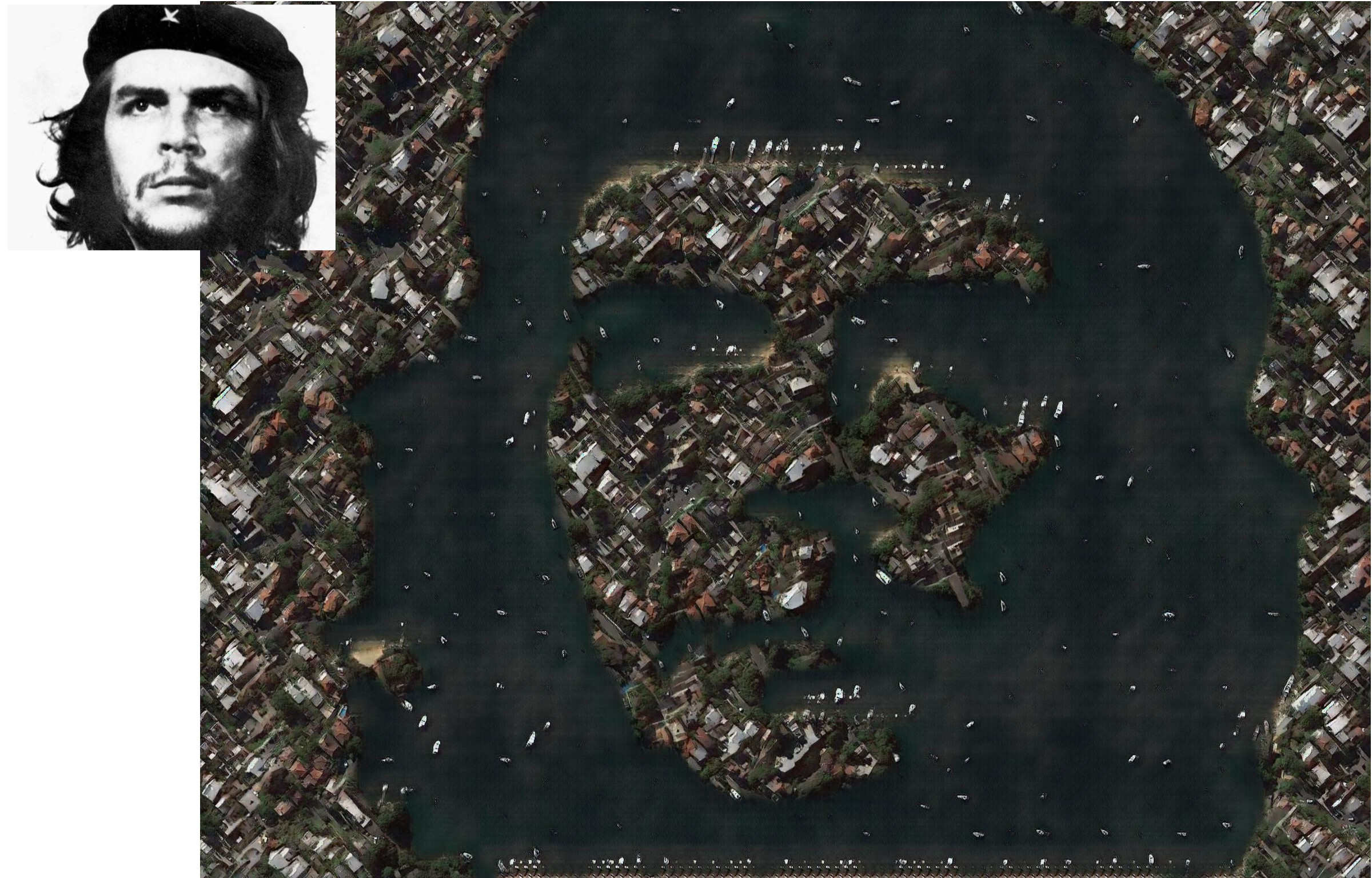}
\caption{Random projections are useful for texture exploration, here the Sydney satellite views. A content image of size 64x48 was projected to global noise channels $Z^g$ of the same size spatially, and local noise $Z^l$ from the prior. The resulting $G(Z)$  generates a 2048x1536 pixel output image. }\label{fig_che}
\end{figure}

A very specific choice of correspondence map will use a map $\phi$ which always outputs a constant value, equivalent to disabling the whole content loss term in the GANosaic loss.
In that case, we introduce a simpler alternative method that can create mosaic images using a PSGAN texture generator. We can directly paint the noise global dimensions $Z^g$ with values related to the pixels of the content image. 
E.g. we can use a random linear projection from pixels (3 channels) to the $d^g$ channels of the global noise tensor, followed by a nonlinearity to keep the values in $[-1,1]$. 
Concretely, given an image $I$ of size $H \times W$ pixels, we can calculate a downscaled version $\hat{I}$ of size $L \times M$ pixels -- the spatial resolution of the latent noise space of $G$.
We can then sample a random matrix $R \in \mathbb{R}^{d^g \times 3}$ and calculate per spatial position $Z^g_{\lambda,\mu} = \sin(R\hat{I}_{\lambda,\mu})$. 
Afterwards we can generate an image $G(Z)$ with the projection $Z^g$ and local noise $Z^l$ from the prior.

While very simple, this approach is useful for exploration of texture manifolds and is very fast to calculate. 
Figure \ref{fig_che} shows an example: the random projection paints the low frequency segments of the content image, while exhibiting a lot of details from the texture manifold. This is useful for creating smoothly morphing videos illustrating random walks in the space of the texture manifolds, while projecting a specific content image, and usually preserving well the low frequencies of the content. Please see an example video of Che Guevara rendered with Sydney textures at \url{youtu.be/4GAFQwE3kLs}.
As a drawback, the random projection mosaics completely ignore the high frequencies of the content image. The random projection output is also more unpredictable: some projections look better than others, but the fast generation speed (0.1s for the 2048x1536 pixel image in Figure \ref{fig_che}) enables exploring many such images as artistic selection process. 

\subsection{Effects of the texture loss}

\begin{figure}
\centering
\subfigure[No $Z^l$ optimisation.]{\includegraphics[width=4cm]{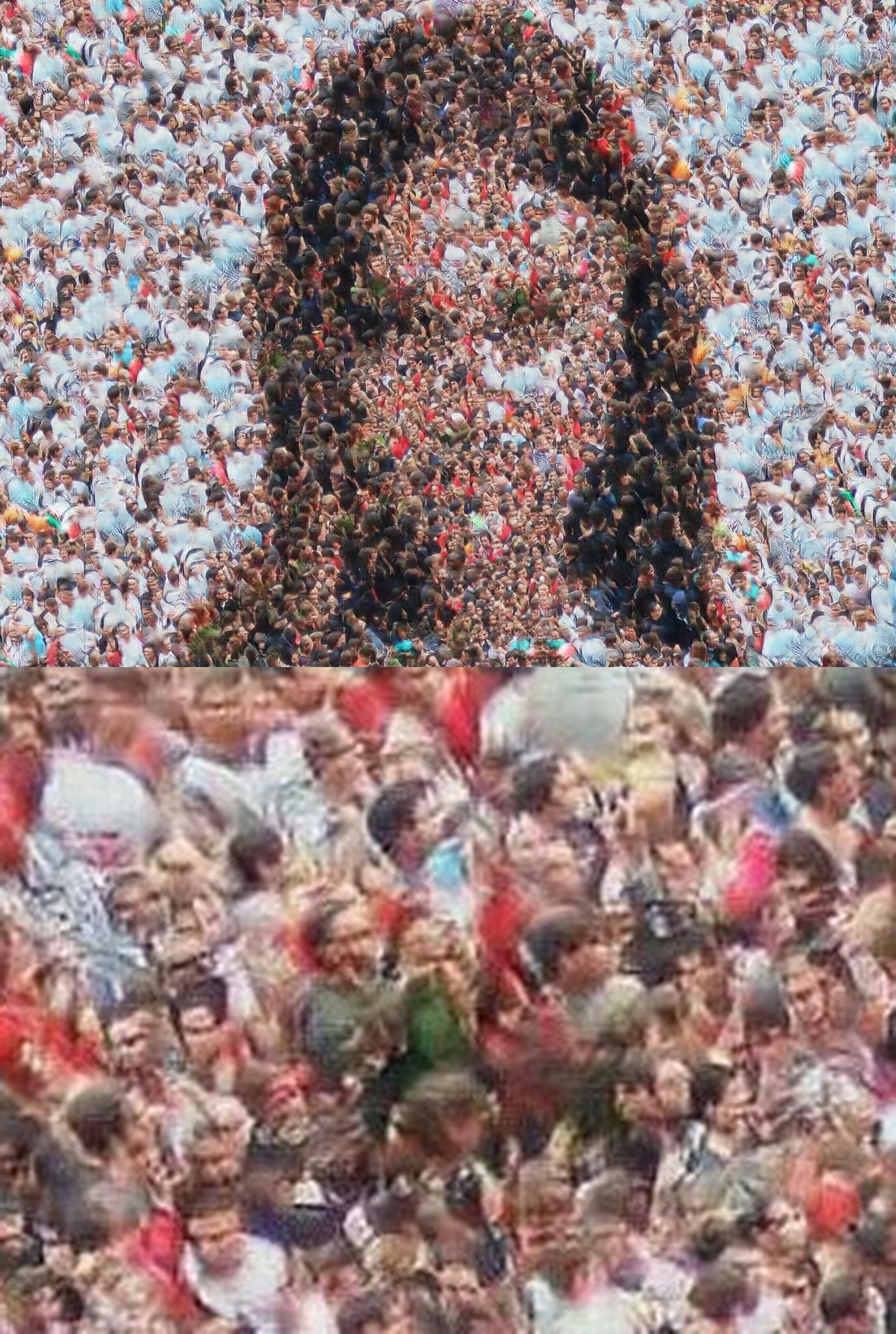} }
\subfigure[No texture loss, $\alpha_l =0$.]{\includegraphics[width=4cm]{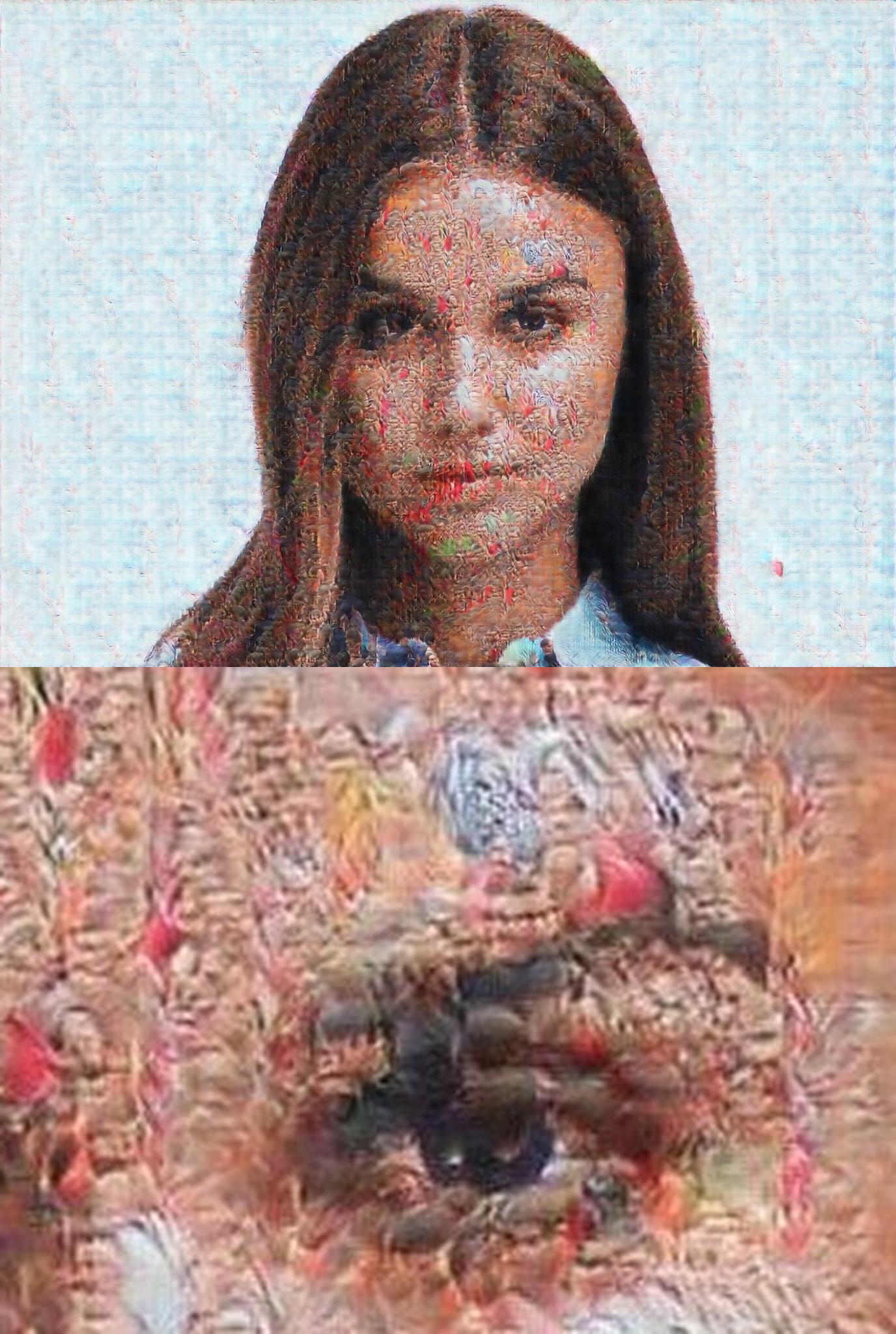} }
\subfigure[Texture loss active, $\alpha_l =5$.]{\includegraphics[width=4cm]{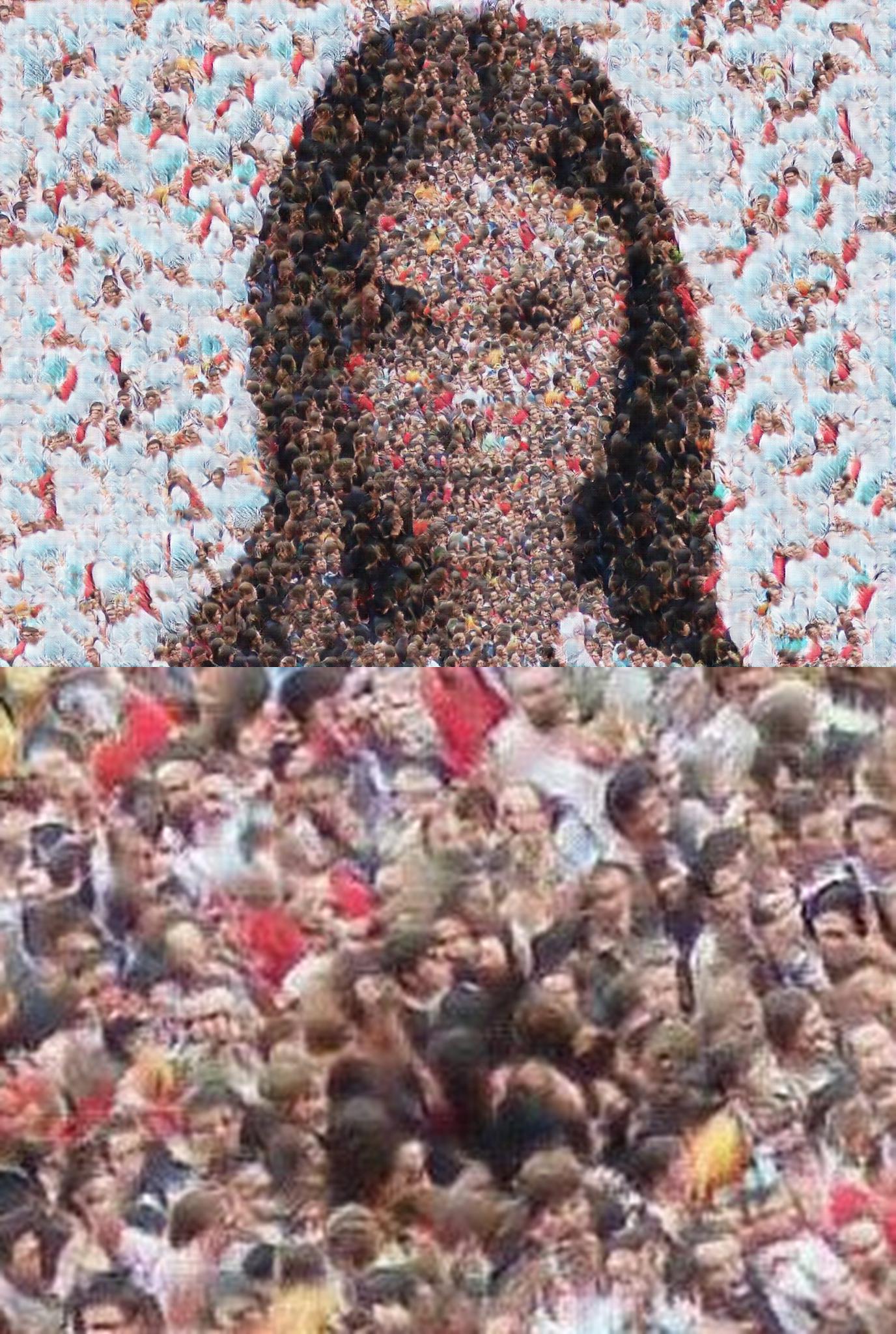}  }
\caption{Different mosaic results, with a zoom-in on the right eye for details. (a) optimizes only the global dimensions and has worse content fit than (b) and (c) which also optimize the local dimensions and so can fit the content better. However, in (b) the lower content loss is also combined with degeneration from the texture manifold: see how the people in the texture become blurry. In (c) the texture loss corrects the degeneration, while still fitting the content better than (a). A texture of many small people in a crowd is used, which emphasizes the visual degeneracy (b) or correct random-like texture appearance (a,c).}\label{fig_regl1}
\end{figure}

Optimizing the local values $Z^l$ together with the global $Z^g$ can lead to lower content loss than tuning $Z^g$ alone and fixing $Z^l$ to a random sample from the prior. However, special care needs to be taken to keep the distribution of the $Z^l$ close to the prior distribution and avoid degeneracy from the texture manifold -- thus the texture loss term $\mathcal{L}_{tex}$ we defined in Section \ref{sec_locreg}.
Figure \ref{fig_regl1} shows graphically the mosaic quality obtained by optimizing and regularizing $Z^l$ by using the texture loss. This term acts as a regularization term that is helpful for preventing degeneration. It also allows GANosaic to reach a better content loss by optimizing both $Z^g$ and $Z^l$, rather than just $Z^g$. 

\begin{figure}[tb]
\centering
\subfigure[Content loss $\mathcal{L}_c$.]{\includegraphics[width=5.cm]{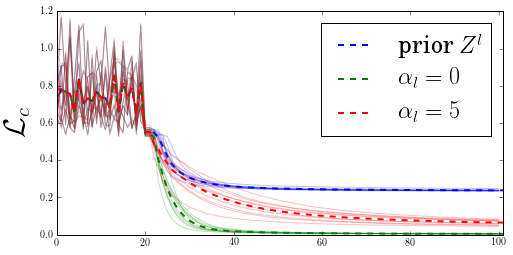} }
\subfigure[$\alpha_l =0$]{\includegraphics[width=3.5cm]{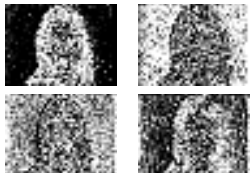} }

\subfigure[Regularization term $\mathcal{L}_{tex}$.]{\includegraphics[width=5.cm]{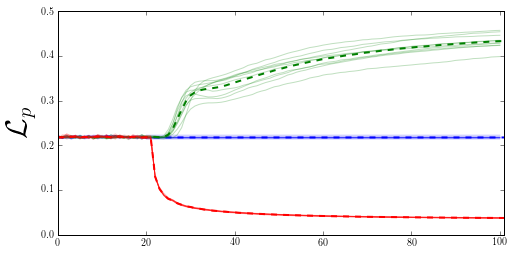} }
\subfigure[$\alpha_l =5$]{\includegraphics[width=3.5cm]{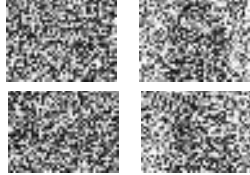} }
\caption{(a,c) show optimization convergence plots: after 20 random projection steps the optimizer further decreases the costs in 80 BFGS gradient steps. We compare not optimizing $Z^l$ (blue), optimizing $Z^l$ and setting $\alpha_l=0$ (green) and regularizing by setting $\alpha_l=5$ (red). The thin lines are 10 different runs, and the dashed line their mean. In (b) we show as images 4 channels of $Z^l$ after optimization with $\alpha_l=0$ -- the values are correlated with the content image, which is different than the random noise prior and can lead to degenerate looking textures in the output mosaic. In (d) we see the effect of $\alpha_l=5$ to make the channels of $Z^l$ closer to the prior.}
\label{fig_regl2}
\end{figure}

Figure \ref{fig_regl2} shows plots of the convergence and emphasizes the optimization behaviour for the 3 different settings from the previous figure. Plot \ref{fig_regl2}(a) shows the values of the content loss when optimizing with the different settings and we see how optimizing $Z^l$ leads to lower content loss. The texture loss is effectively lowered when we set $\alpha_l =5$, see \ref{fig_regl2}(c). For intuition how the tensors $Z^l$ look, we display as images 4 channels from the  $Z^l$ tensor. In Figure \ref{fig_regl2}(b) they are correlated with the content image and deviate from the prior. Such values of $Z^l$ can lead to the degeneracy displayed in Figure \ref{fig_regl1}(b). In \ref{fig_regl2}(d) the effect of the term $\mathcal{L}_{tex}$ de-correlates $Z^l$ from the content image and makes them locally consistent with the uniform prior, which corresponds to the mosaic from Figure \ref{fig_regl1}(c).

\section{Discussion}
\label{sec_discuss}
This section contains a short discussion of the properties of GANosaic mosaics.
Traditionally, photomosaic algorithms utilized large image datasets~\cite{photomosaic,JIM}. Texture rendering is usually done with a single texture~\cite{EfrosQ,GatysEB15a}. 
In contrast, GANosaic can use rich texture manifolds as style representations and allows the generation via convolutional neural networks of large mosaics smoothly rendering any content image. The generation of texture mosaics is achieved by optimization in the latent noise space of a PSGAN texture model. 
An application of the GANosaic can be consumer-facing apps that allow quick rendering of user content (images and video), using pre-trained texture models.
Another more professional use case can be graphical design: the artist can use PSGAN and GANosaic as tools in the creative process: 
\begin{itemize}
\item select carefully a set of texture examples
\item train a PSGAN texture model on them
\item use the texture model on any content image to create high resolution art, suitable for posters.
\end{itemize}
As a practical tip, it is recommended to the user to regard the GANosaic process not as fully automatic, but semi-automatic -- the artist can explore the results of different initializations and iterations of the optimizer,  and select mosaics that show textures with the ``right'' aesthetic look.

The fast runtime and the ability to handle arbitrarily large output image resolutions comes from the spatial GAN architecture~\cite{SGAN2016} used in the texture prior of GANosaic. The generation of $G(Z)$ can be easily split by calculating $G$ on separate subtensors of $Z$ and combining the results. This property applies also to our optimization framework, since all the loss function terms can be calculated and aggregated in spatial chunks. Thus, mosaics of very high resolution, practically unlimited by memory (only by storage) can be made. The computation time scales linearly with the number of output pixels.

In order to minimize the mosaic loss in Equation \ref{eq_loss}, different spatial regions of the input noise tensor $Z^g$ will converge to different values and thus the image output from the generator will be a mosaic of different texture processes. While this works well for the low frequencies in the content image, there is a limit to how high the frequencies in the output mosaic can be. The receptive field of the generator model determines the highest possible frequencies we can obtain in the mosaics by setting the generator input.   
Optimizing both the local and global dimensions improves the texture mosaic resolution and allows to paint finer details better fitting the content, but there is always some limit how high the mosaic image frequencies can go and what level of pixel detail is achievable.


Downscaling in the map $\phi$ can improve the image quality of the output mosaic (see Figure \ref{fig_dsample}) -- it acts like an averaging filter that removes the high frequencies from the content image. Thus, the content loss depends on the lower frequencies, and the high frequencies will be determined by the texture manifold.
The choice of content image and its pixel size matter as well. Large content images have lower frequencies and are easier for mosaics.  
In practice, we can always upscale the input image to obtain larger mosaics with lower frequencies. Using a smaller texture brush relative to the image leads to a lower content loss, analogous to a large photomosaic with small tiles. 
On the other hand, smaller content images imply larger textures relative to them ( e.g. the works of Archimboldo) and this is a more challenging case for mosaic optimization. 



\begin{figure}
\centering
\subfigure[Texture style image.]{\includegraphics[width=4.75 cm]{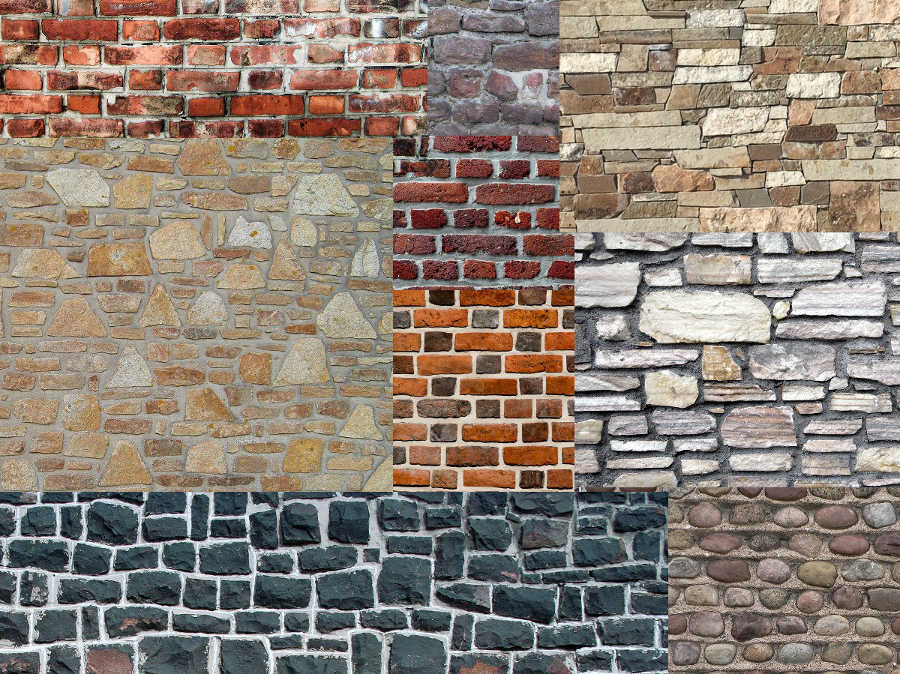} }
\subfigure[Style transfer image.]{\includegraphics[width=4.75 cm]{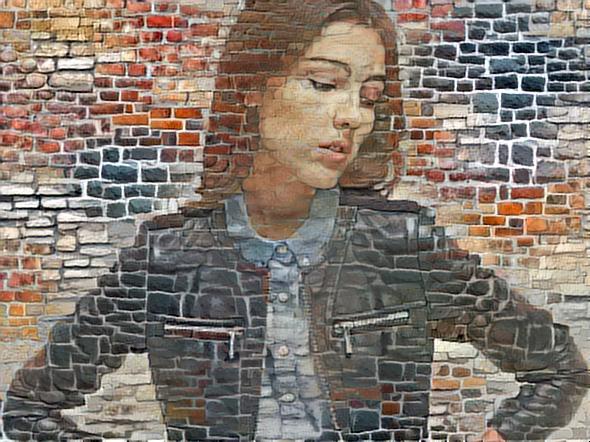} }
\vspace{-0.3cm}
\caption{Using the (composite) style image (a), and a human photography as content (original shown in Figure 1) the neural style transfer method produces the image shown in (b). A detailed look at that output image reveals interesting artifacts. The jacket segment is a good stone mosaic, but the face segment has issues: it is rendered too similar to the content pixels and lacks the stone texture. }\label{fig_gatys}
\end{figure}

The approach of GANosaic to locally fit the ``right" texture is different than the neural art style approach~\cite{GatysEB15a}. Figure \ref{fig_gatys} shows an example of neural style transfer when using a style image with many textures. The style descriptor consists of the feature correlations marginalized over the spatial extent of the image, and 
style transfer will try to transfer the full distribution of the style image to the content -- thus the mixed stone background. 
The pixel space optimization can preserve the high frequency details of the content image, but in some regions (e.g. the face in Figure \ref{fig_gatys} b)) it looks like merely painting directly the content, rather than trying to represent it with textures.

In contrast, GANosaic works by modifying \textit{only} the input noise tensor $Z$ to the generator network $G$. This network acts as a regularizer that outputs images $G(Z)$ that are close to the underlying PSGAN texture manifold. GANosaic does not optimize directly the pixels of the output mosaic image, but only modifies $Z$ as input of $G$. This is a more constrained optimization problem since the noise space $Z$ has much lower spatial size than the image $I$. E.g. with $G$ with 5 layers an image with 1024x1024 pixels will be generated by a tensor $Z$ of size 32x32 spatial positions.

In a related work,~\cite{DumoulinSK16} explored the blending of multiple styles by parametrically mixing their statistical descriptors, but the result is a style averaging globally the properties of the input styles, rather than painting locally with different styles.  
In contrast, the GANosaic (e.g. as shown in Figure \ref{fig_dsample} with the same texture style) can focus on different modes from the texture distribution that well approximates the content image locally, and the regularization ensures that each local region is on the texture manifold.

\newpage
\section*{Acknowledgments}
The authors would like to thank Roland Vollgraf for useful discussions and feedback.
\bibliography{bibi}
\bibliographystyle{plain}

\end{document}